\documentclass[preprint,10pt]{elsarticle}

\usepackage[utf8]{inputenc} % allow utf-8 input
\usepackage[T1]{fontenc}    % use 8-bit T1 fonts
\usepackage{hyperref}       % hyperlinks
\usepackage{url}            % simple URL typesetting
\usepackage{booktabs}       % professional-quality tables
\usepackage{amsfonts}       % blackboard math symbols
\usepackage{nicefrac}       % compact symbols for 1/2, etc.
\usepackage{microtype}      % microtypography

\usepackage{amssymb}
\usepackage{bm}
\usepackage{amsmath}
\usepackage{graphicx}
\usepackage{epstopdf}
\usepackage{subfigure}
\usepackage{fullpage}

\begin{document}
\begin{frontmatter}

\title{Thick-Net: Parallel Network Structure for Sequential Modeling}
\author[lab1]{Yu-Xuan Li}
\ead{lyxjoooshua@gmail.com}
\author[lab1]{Jin-Yuan Liu}
\ead{jinyuanjames@gmail.com}
\author[lab2]{Liang Li\corref{cor1}}
\ead{plum\_liliang@uestc.edu.cn, plum.liliang@gmail.com}
\author[lab3]{Xiang Guan}
\ead{duochuan.gx@gmail.com}

\address[lab1]{Glasgow College, University of
               Electronic Science and Technology of China, 611731,
               Chengdu, P.R. China}

\address[lab2]{School of Mathematical Sciences, University of
               Electronic Science and Technology of China, 611731,
               Chengdu, P.R. China}

\address[lab3]{School of Computer Science and Engineering, University of
               Electronic Science and Technology of China, 611731,
               Chengdu, P.R. China}

\cortext[cor1]{Corresponding author}

\begin{abstract}
Recurrent neural networks have been widely used in sequence learning tasks. In previous studies, the performance of the model has always been improved by either wider or deeper structures. However, the former becomes more prone to overfitting, while the latter is difficult to optimize. In this paper, we propose a simple new model named Thick-Net, by expanding the network from another dimension: “thickness”. Multiple parallel values are obtained via more sets of parameters in each hidden state, and the maximum value is selected as the final output among parallel intermediate outputs. Notably, Thick-Net can efficiently avoid overfitting, and is easier to optimize than the vanilla structures due to the large dropout affiliated with it. Our model is evaluated on four sequential tasks including adding problem, permuted sequential MNIST, text classification and language modeling. The results of these tasks demonstrate that our model can not only improve accuracy with faster convergence but also facilitate a better generalization ability.
\end{abstract}


\end{frontmatter}

\section{Introduction}
\label{submission}
With the availability of large-scale datasets, high-capacity of various neural networks and powerful computational technology and devices, numerous challenging problems in sequential learning tasks have been solved by employing artificial neural networks. An artificial neural network is an interconnected assembly of nodes (artificial neurons), inspired by a simplification of neurons in animal brains \cite{Gurney1997An}. To enable neural networks to extract richer information and learn better features, increasing network width or depth are considered to be top two options \cite{gilboa2019wider}.

According to the Universal-Approximation Theorem proposed by Cybenko \cite{cybenko:mcss} and Hornik \cite{journals/nn/Hornik91}, one or more layers can universally approximate any continuous functions on compact subsets of $\mathbb{R}^{n}$ when the width of the network is sufficiently large (large number of nodes in one layer). The hypothesis space is the set of all functions that returned by a network, and the functions with their internal parameters can be represented by the interconnection between nodes. As the hypothesis space of a network grows, the wider network can therefore learn richer structures. Theoretically, a sufficiently wide network is able to eventually memorize the corresponding output for every possible input, but there does not exist every possible inputs to train with in practical applications. Besides, more difficulties may occur when using an extremely wide network. Despite the strong memorization, the network will become more prone to overfitting and its generalization ability tends to be relatively poor.

Inside the neural networks, the size of the hypothesis space is determined by the total number of nodes. For a fixed number of nodes, there is always a basic trade-off between its width and depth. Instead of increasing width in one layer, using networks that contains many layers with a small number of nodes per layer can be an alternative \cite{Eldan2015The}. This type of nodes layout can be revealed in various networks going from 7 layers (AlexNet \cite{conf/nips/KrizhevskySH12}) to even thousands of layers (ResNet \cite{conf/nips/HeGXLHB17}).

It is noteworthy that this trend, i.e., increased depth, widely exists in vision-related tasks and convolutional networks. It can be interpreted as multiple layers that can extract features at various levels of abstraction. Whereas for RNNs, they obtain sequence representation by recursively updating hidden units using linear transformations and nonlinear activation functions. For each step, one single cell of RNNs can only extract current input and previous hypothesis spaces. Owing to the structural limitations of RNNs, only previous features rather than various levels of abstraction are extracted repeatedly when the networks become deeper. At the same time, deeper networks make the optimization more difficult. This is the reason why RNNs are generally not as deep as ResNets with thousands of layers. Therefore, relatively shallow networks are usually used in RNNs for several tasks including text classification \cite{Bahdanau2016} and language modeling \cite{conf/iclr/MerityKS18}.

Pervious researchers have the tendency to enlarge the width or depth of the networks. However, different disadvantages may appear when applying them into practice. For instance, wider networks can easily result in overfitting and increasing generalization error, while deeper networks lead to a more difficult optimization.

In this paper, we increase the parameters in one RNN cell in another smaller dimension without increasing the hidden state size (enlarging the hypothesis space), and simply define it as “Thick-Net”. A cluster of values are obtained in each node of RNN by different sets of parameters, among which only the maximum values of each set are fed into the next node affiliated with the dropout for the rest of values. This maximization operation is a form of non-linear down-sampling. In order to avoid the gradient vanishing caused by the selection of the maximum, we apply batch normalization before the non-linear activation function \cite{conf/icml/IoffeS15}. %In addition, ReLu activation function is explicitly applied.

We summarize our contributions in this work as follows:
\begin{itemize}
\item We present a novel RNN structure allowing more parameters in one single node which can be filtered through maximization operation. In other words, our model can learn richer structures by increasing the thickness of each node instead of increasing the number of layers.
\item The maximization operation applied in our paper reduces the dimension of hypothesis space which can also be understood as the down-sampling, and hence our model avoids the overfitting appearing in wider networks.
\item Although the proposed maximization operation leads to a higher dropout rate, gradient information is still preserved through the back propagation. All the parameters are optimized in each training step which makes the model easier to optimize.
\end{itemize}

In our experiment, we test the effectiveness of our approach on four sequence modeling tasks: the adding problem, permuted sequential MNIST, text classification and language modeling. We run extensive comparisons with multiple baseline models and achieve state-of-the-art performance. Experiments show that our proposed Thick-Net is easier to optimize and better in generalization.

The rest of the paper is organized as follows. The related work is reviewed in Section $2$. The maximization operation, Thick-Net and its embedding in Long Short-Term Memory (LSTM) \cite{lstm} are described in Section $3$. The performance of Thick-Net is evaluated in Section $4$. The conclusion and future work are described in Section $5$.

\section{Related Work}
There is a large body of work focusing on sequence modeling tasks by applying various neural networks. So far, the RNN and its variants are more suited for tasks involving sequential or temporal data, and the most widely used ones are LSTM \cite{lstm} and Gated Recurrent Units (GRU) \cite{conf/emnlp/ChoMGBBSB14}. These two typical variants have been introduced to control gradient vanishing and explosion which are commonly found in long sequence RNN tasks \cite{BengioSimardFrasconi94}. The gating mechanism in these two networks controls which part of the present inputs and previous state memory are used to update the current activation function and current state.

Recent efforts have been introduced to continuously improve the performance on dealing with long sequence, including acceleration of convergence during training process and optimization of the internal parameters. Adding an extensional state update gate can allow skip state update and thereby reducing the number of sequential operations, where computation in RNN may or may not be executed in each time step \cite{conf/iclr/CamposJNTC18}.

In addition to changing the structure of the recurrent neural networks, works on increasing the number of nodes in width and depth have also been proposed. A simple technique called parallel cells to enhance the learning ability of RNNs has been proposed \cite{zhugoing} where in each layer there are multiple small RNN cells rather than one single large cell. Zhen et al. proposed the Tensorized LSTM \cite{conf/nips/HeGXLHB17} in which by increasing the tensor size and delaying the output, the network can be widened efficiently and deepened implicitly respectively. Fast-Slow RNN (FS-RNN) \cite{mujika2017fastslow}, a novel recurrent neural network (RNN) architecture, has been introduced which combines the strengths of both multiscale RNNs and deep transition RNNs. Wide linear models and deep neural networks are also jointly trained to combine their advantages of memorization and generalization in recommendation system tasks \cite{Cheng_2016}.

There are other approaches offering interesting trade-off for increasing parameters without increasing hidden state size. Recurrent highway network extends the LSTM architecture to allow step-to-step transition depths larger than one, which controls the recurrent state size to remain still. Maxout non-linearity \cite{goodfellow2013maxout} has always been considered as an activation function to replace often-used ReLU or sigmoid functions in feed forward networks. Using a Maxout unit between two layers allows us to train multiple sets of parameters and then select the set with the maximum activated value. The saturating non-linearity (tanh activation functions) in one LSTM cell can be modified by non-saturating activation functions (Maxout units) without causing the instability of the model \cite{Gulcehre_2014,7178842}.
%-------------------------------------------------------------------------
\section{The Proposed Thick-Net}
%-------------------------------------------------------------------------
In this section, we articulate our Thick-Net in detail. The maximization operation is firstly presented in $3.1$. Thick-Net and its embedding in LSTM architecture "Thick-LSTM" are then described in $3.2$ and $3.3$, respectively.

\subsection{Maximization Operation}
Firstly, we introduce a downsampling method called maximization operation, which is the premise of Thick-Net.
Maximization operation takes $n$ vectors, ${\bm{v_1},\bm{v_2},\cdots ,\bm{v_n}}$ as inputs, $n\in \mathbb{Z}^{+}$ and $\bm{v_i}\in \mathbb{R}^{m}$. Then the maximization operation is defined as:
\begin{equation}
 \begin{split}
 &MAX: \mathbb{R}^{m\times n}\rightarrow \mathbb{R}^{m},\\
 &MAX(\bm{v_1},\bm{v_2},\cdots ,\bm{v_n}) \\
 = &\{max(v_{11},\cdots ,v_{1n}),\\
 &\cdots,\\
 &max(v_{m1},\cdots ,v_{mn})\}\\
 \end{split}
 \end{equation}
where $v_{ij}\in \mathbb{R}$ represents the $j-th$ element of $\bm{v_i}$.
The output value of $MAX(*,*,\cdots ,*)$ is determined by the selection of the maximum from corresponding values of a set of inputs.

This maximization operation can be considered as the an extension of Maxout units \cite{goodfellow2013maxout} or max pooling, which shows a similarity to select the maximum among a cluster of values.
%-------------------------------------------------------------------------
\subsection{Thick-Net}
Unlike traditional neural networks that map input to a point in a high-dimensional space, our proposed Thick-Net maps inputs to a cluster of points in space. The output is obtained by down-sampling these points with the maximization operation described in $3.1$. More features are acquired through multiple points, followed by a down-sampling that avoids overfitting and improves generalization ability by controlling the size of the hypothesis space.

It performs multiple linear transforms and takes the maximization operation of all linear transforms in each hidden state as a role of dropout. Thick-Net is an expansion of matrix multiplication which can be applied in every linear transforms of the networks, while maxout unit performs only as an alternative of non-linearity.

For the input $\bm{x}\in \mathbb{R}^{r}$, we use $n$ matrices $W^{(i)}\in \mathbb{R}^{m\times r}$ to linearly transform it to obtain an output of length $m$ and thickness $n$. Then down-sampling function along the thickness direction is performed by maximization operation:

\begin{equation}
 \begin{split}
 &MAX(W^{(1)}\cdot \bm{x},W^{(2)}\cdot \bm{x},\cdots ,W^{(n)}\cdot \bm{x}) \\
 = &\{max(\sum_{k=1}^rw^{(1)}_{1k}x_k, \sum_{k=1}^rw^{(2)}_{1k}x_k,\cdots , \sum_{k=1}^rw^{(n)}_{1k}x_k),\\
 &\cdots,\\
 &max(\sum_{k=1}^rw^{(1)}_{mk}x_k, \sum_{k=1}^rw^{(2)}_{mk}x_k,\cdots , \sum_{k=1}^rw^{(n)}_{mk}x_k)\}\\
 \end{split}
 \end{equation}
where $\cdot$ denotes the matrix multiplication, $w^{(i)}_{jk}\in \mathbb{R}$ represents the $j$-th row and the $k$-th column of $W^{(i)}$, and $x_{k}\in \mathbb{R}$ represents the $k$-th element of $\bm{x}$.

\begin{figure}[t]
\vskip 0.2in
\begin{center}
\centerline{\includegraphics[width=0.6\columnwidth]{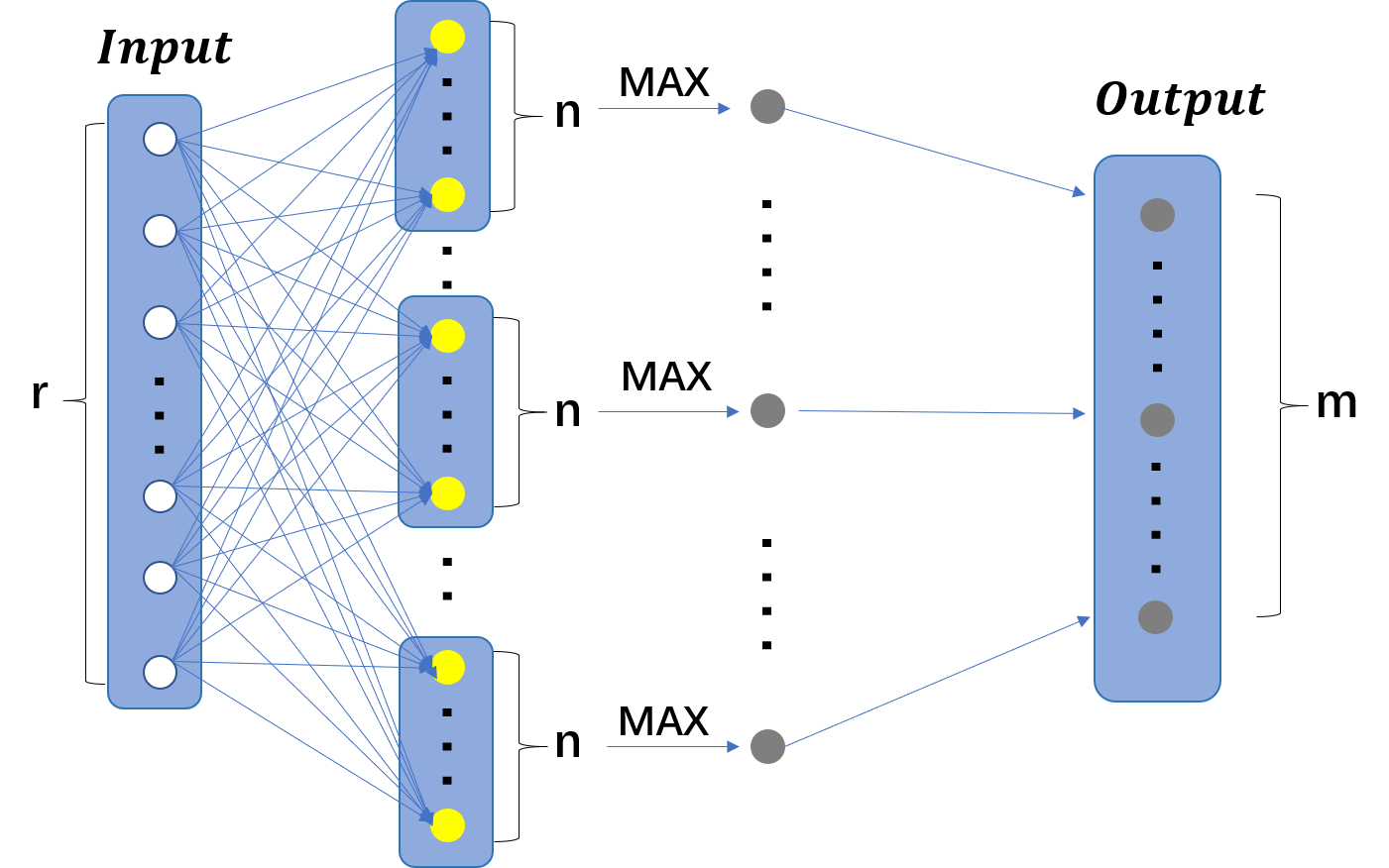}}
\caption{Structure of Thick-Net. The $r$-dimensional vector is considered as the input, which generates $m$ nodes through $n$ linear transformations. Maximization operation is applied along the thickness direction to form the $m$-dimensional vector as the output}
\label{model}
\end{center}
\vskip -0.2in
\end{figure}

As shown in Figure ~\ref{model}, for the input $\bm{x}$, $m$ nodes are obtained through $n$ sets of linear transformations, each node contains $n$ values. Furthermore, the maximum value in each node is selected, and the rest is dropped out. Finally, the m-dimensional output vector is obtained by maximization operation mentioned in $3.1$.

Owing to this operation, our proposed Thick-Net have several superiorities. This structure does not only increase the size of the hypothesis spaces, but also introduces a large dropout rate through the maximization operation. Additionally, Thick-Net avoids the overfitting of the training dataset happening in the wide neural network.

Even though more parameters are introduced, all the parameters are set in parallel on a one-layer network. Notably, there is no complicated chain-like derivation process during backpropagation. Moreover, the above dropout only drops partial values in a batch of data, but reserves the gradient information of the parameter. This enables the model easier to optimize compared with the neural network which expands the model in depth.
%-------------------------------------------------------------------------
\subsection{Thick-LSTM}
In this part, we embed our proposed Thick-Net into LSTM architecture for sequential learning tasks. For these tasks, traditional RNNs update the hidden state over time with a fixed linear transformation. To extract features from more complex and variable inputs of the sequence modeling tasks, we update the hidden state of each step by introducing Thick-Net. This structure allows each hidden state to be derived from a set of linear transformations to learn richer structures.

\begin{figure}[t]
\centering
\subfigure[LSTM]{
\begin{minipage}[t]{0.29\linewidth}
\centering
\includegraphics[width=2.4in,trim=0 15 0 43,clip]{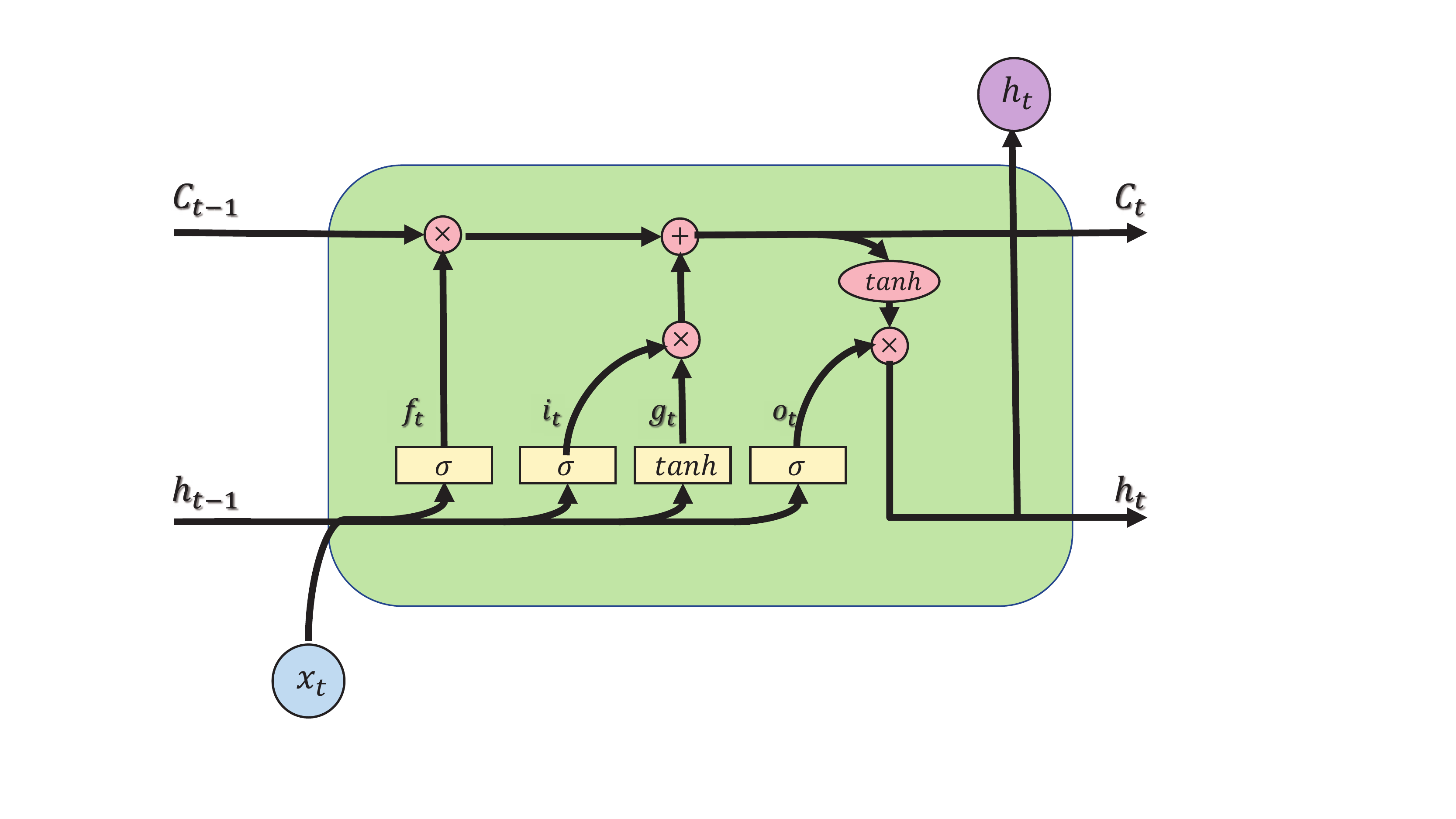}
\label{fig:LSTM}
\end{minipage}%
}%
\subfigure[Maxout LSTM]{
\begin{minipage}[t]{0.29\linewidth}
\centering
\includegraphics[width=2.4in,trim=0 15 0 43,clip]{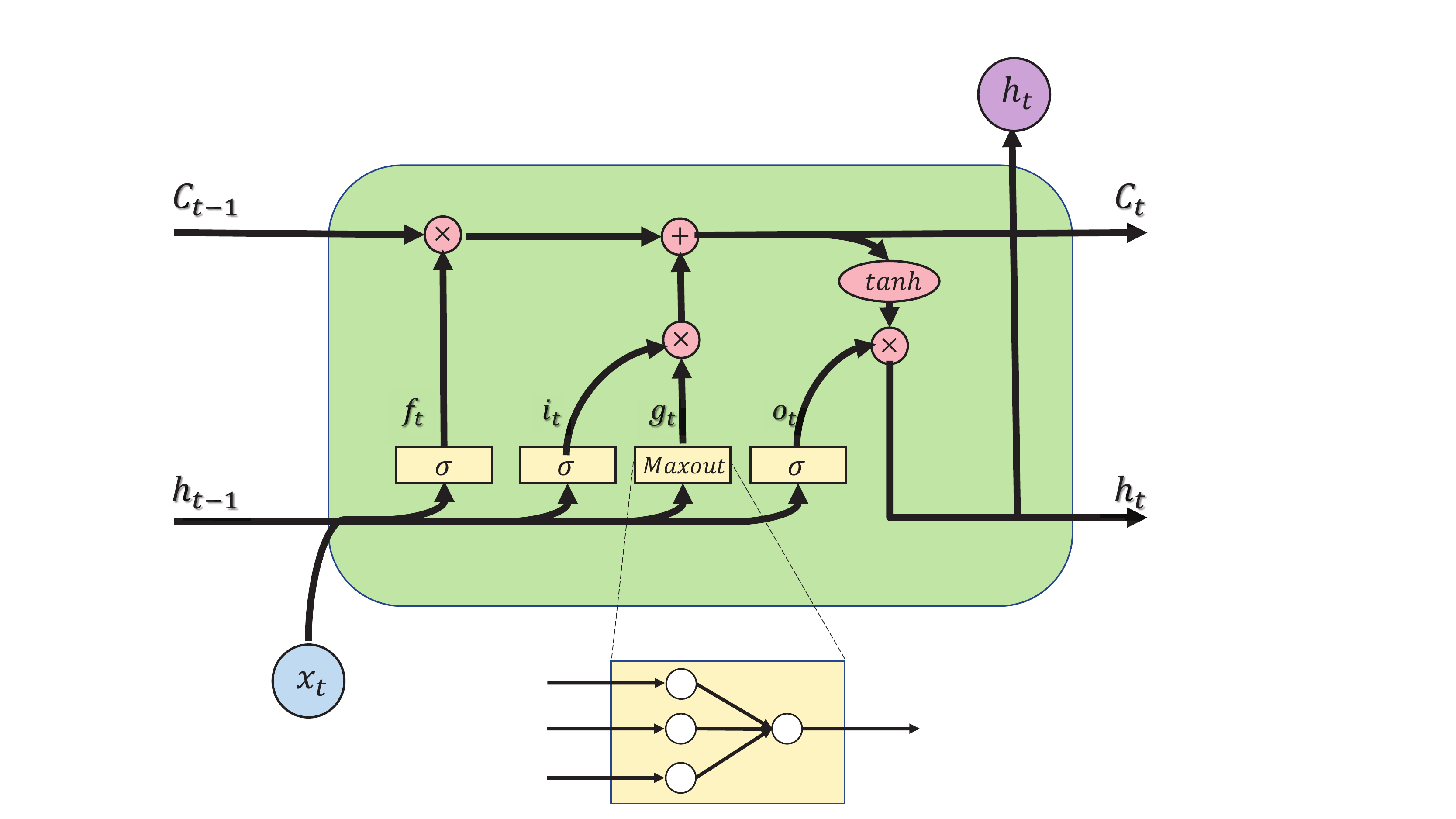}
\label{fig:MaxoutLSTM}
\end{minipage}%
}%
\subfigure[Thick-Net]{
\begin{minipage}[t]{0.29\linewidth}
\centering
\includegraphics[width=2.4in,trim=0 15 0 43,clip]{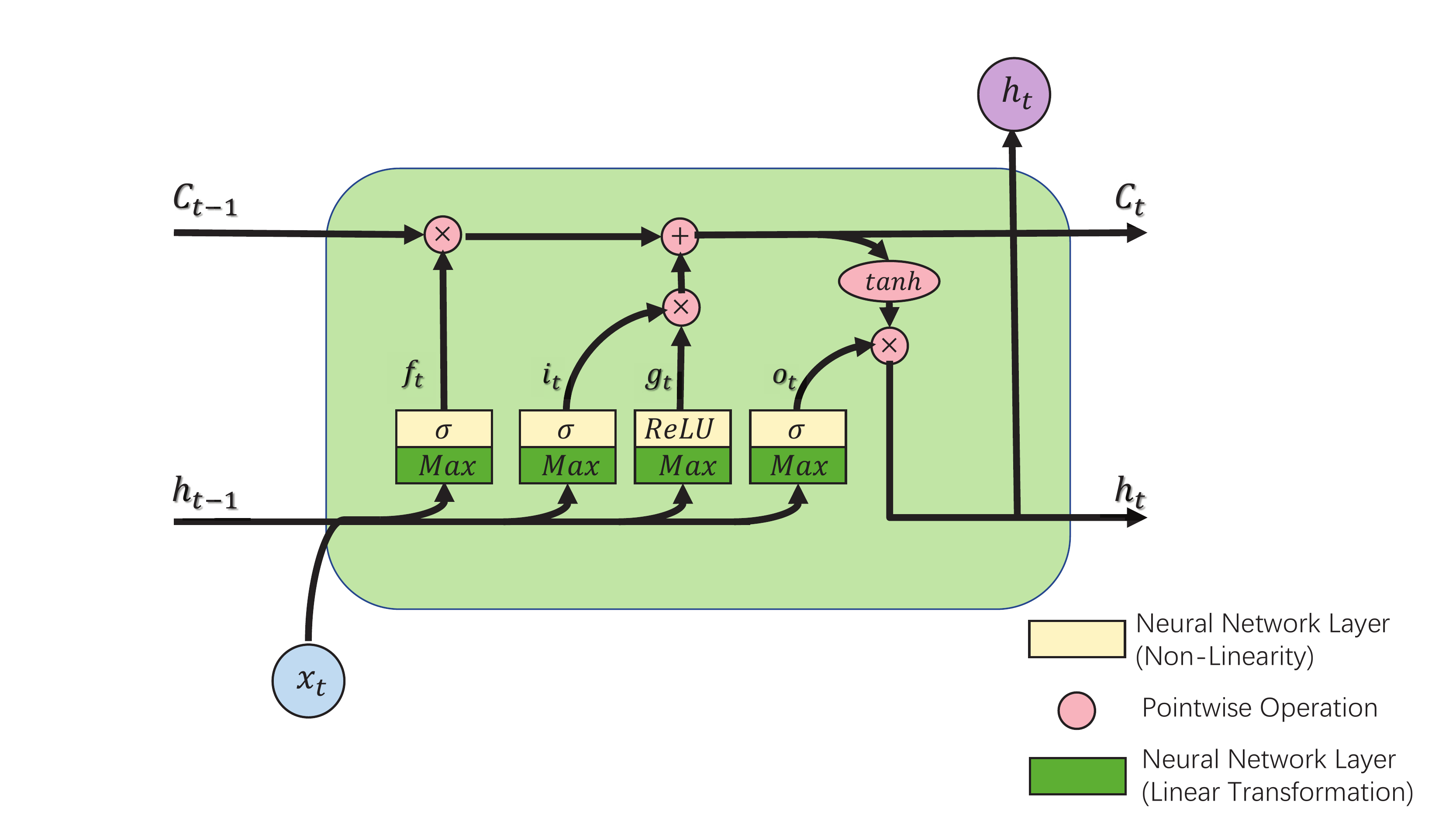}
\label{fig:ThickNet}
\end{minipage}%
}%
\centering
\caption{Architectures of LSTM, Maxout LSTM and Thick-LSTM networks in one block}
\label{Threemodels}
\end{figure}

In order to avoid gradient explosion and vanishing of traditional RNN, we apply our Thick-Net to a special recurrent neural network called long short-term memory model (LSTM). The architectures of LSTM, Maxout and our proposed Thick-LSTM are shown in Figure \ref{Threemodels}. We can see that in a Maxout unit one extends the parameters of the non-linear activations and then extracts the only set of parameters with the maximum activation value, while in a Thick-Net unit we multiply the sets of linear transformation and then select the maximum value to pass towards non-linear activation functions.

The standard architecture of LSTM applies a range of repeated modules for each time step as in a RNN, and these steps in LSTM are controlled by a memory cell containing four components: the forget gate $f_t$, the input gate $i_t$, the output gate $o_t$, and the memory cell $c_t$. The gating mechanism can determine which feature gets stored or forgot from the memory based on the current input and cell state. We replace the linear transformation in each gate with our nonlinear transformation Thick-Net. For node in each gate of this Thick-LSTM, multiple parallel results are obtained by different sets of parameters, in which only the maximum value is passed to the next node, and the rest is dropped out. In order to avoid the gradient vanishing caused by the selection of the maximum, batch normalization is applied before the non-linear activation function. The Thick-LSTM transition functions are described as follows:

\[\tilde{i_t}=\sigma (MAX(W_{ih0}\cdot h_{t-1}, \cdots,W_{ihn}\cdot h_{t-1})+MAX(W_{ix0}\cdot x_t, \cdots,W_{ixn}\cdot x_t)+b_{i}),\]%function
\[\tilde{f_t}=\sigma (MAX(W_{fh0}\cdot h_{t-1}, \cdots,W_{fhn}\cdot h_{t-1})+MAX(W_{fx0}\cdot x_t, \cdots,W_{fxn}\cdot x_t)+b_{f}),\]%function
\[\tilde{o_t}=\sigma (MAX(W_{oh0}\cdot h_{t-1}, \cdots,W_{ohn}\cdot h_{t-1})+MAX(W_{ox0}\cdot x_t, \cdots,W_{oxn}\cdot x_t)+b_{o}),\]%function
\[\tilde{g_t}=\sigma (MAX(W_{gh0}\cdot h_{t-1}, \cdots,W_{ghn}\cdot h_{t-1})+MAX(W_{gx0}\cdot x_t, \cdots,W_{gxn}\cdot x_t)+b_{g}),\]%function
\begin{equation}
c_t=\sigma (BN(\tilde{f_t}))\circ c_{t-1}+\sigma (BN(\tilde{i_t}))\circ ReLu (BN(\tilde{g_t})),\\
\end{equation}
\begin{equation}
h_t=BN(\tilde{o_t})\circ tanh(c_t).\\
\end{equation}

Where $tanh$ is the hyperbolic tangent function which can map value in $[-1,1]$, the $\sigma$ denotes logistic sigmoid function where the output value is into $[0,1]$, and ReLu is the activation function. The $\circ$ denotes the elementwise multiplication. $n\in N$ denotes the thickness of each node. $BN(*)$ represents the batch-normalizing transform \cite{Ioffe2015Batch}, i.e., $BN(\mathbf{x}) = \frac{x- E(x)}{\sqrt{Var(x)}}$, where $E$ denotes the mean and $Var$ denotes the variance.

We assign these values from one of the $n$ transformations of current input $x_t$ and previous hidden state $h_{t-1}$ respectively. It increases the thickness of each node instead of increasing the number of nodes in width and depth. According to gate mechanism, $f_t$ is the function to determine how many features from the previous memory state should be forgot. On the contrary, $i_t$ is the function to determine to what extent the new feature should be stored in the current memory cell. After using $g_t$ to generate the temporary value, we use $g_t$ and the preceding memory cell $c_{t-1}$ to combine with input gate $i_t$ and forget gate $f_t$ respectively to get the current memory cell $c_t$. $o_t$ is to determine the output influenced by current memory cell. Moreover, we use $o_t$ multiplying updated memory cell $c_t$ to generate the current hidden state $h_t$.

Intuitively, for each step of Thick-net hidden state, more parameters are involved in updating, and the maxpooling technique enables us to choose and take the initiative to dropout. Therefore, the proposed Thick-Net can achieve a stronger generalization capability for diverse inputs in sequence modeling tasks.
%-------------------------------------------------------------------------
\section{Experiments}
In this section, we evaluate our proposed Thick-Net on four sequential tasks: the adding problem, text classification, permuted sequential MNIST and language modeling. The experiment results are also compared with the results of several state-of-the-art models.
\subsection{Adding Problem}
%\begin{figure}[t]
%\begin{center}
%  \centerline{\includegraphics[width=\columnwidth,trim=0 78 0 90,clip]{ThickNet1.pdf}}%]{ThickNet1.pdf}}

%\caption{The figure illustrates how loss function of MSE varies with the iterations of training. x-axis represents the training step and y-axis represents the loss function of MSE. (a)Results of Thick-Net with different operation functions. (b)Results of Thick-Net with different thickness $n$.}
%\label{thickness}
%\end{center}
%\vskip -0.2in
%\end{figure}

%\end{figure}

\begin{figure}[t]
\centering
\subfigure[Different Operation Functions]{
\begin{minipage}[t]{0.25\linewidth}
\centering
\includegraphics[width=1.4in,trim=0 15 0 43,clip]{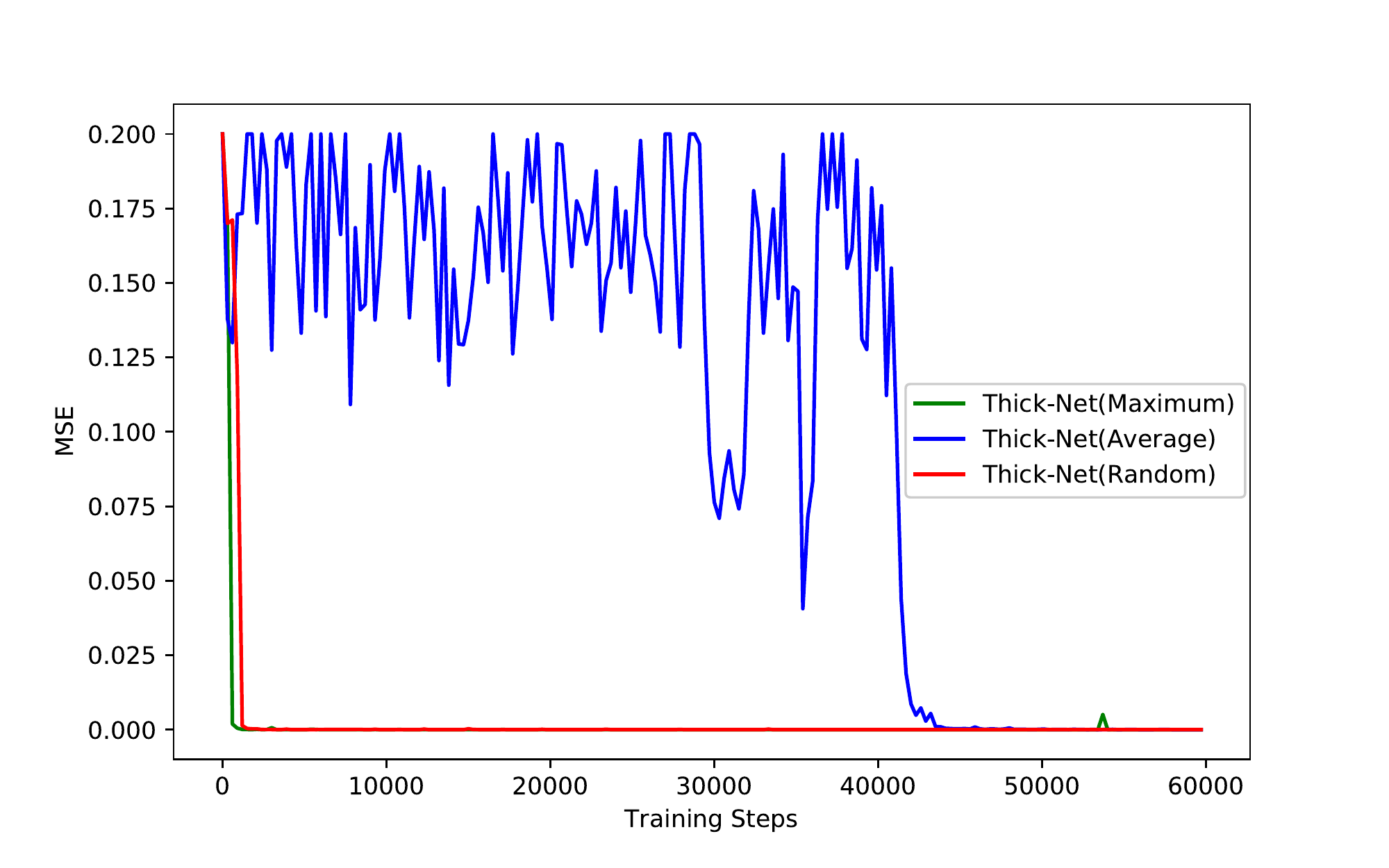}
\label{fig:Different Operation Functions}
\end{minipage}%
}%
\subfigure[Different Thickness]{
\begin{minipage}[t]{0.25\linewidth}
\centering
\includegraphics[width=1.4in,trim=0 15 0 43,clip]{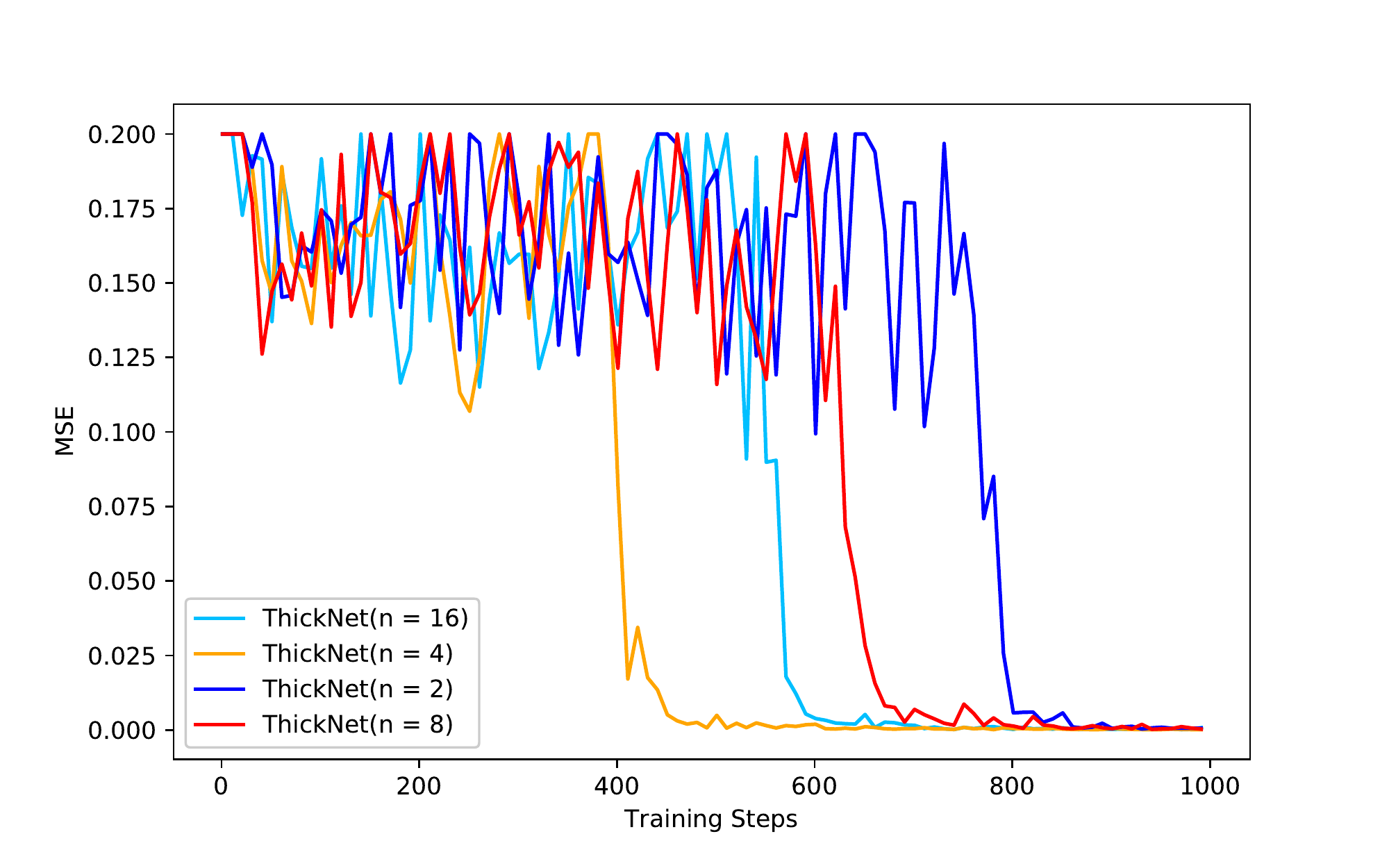}
\label{fig:Different Thickness}
\end{minipage}%
}%
\subfigure[T=100]{
\begin{minipage}[t]{0.25\linewidth}
\centering
\includegraphics[width=1.4in,trim=0 15 0 43,clip]{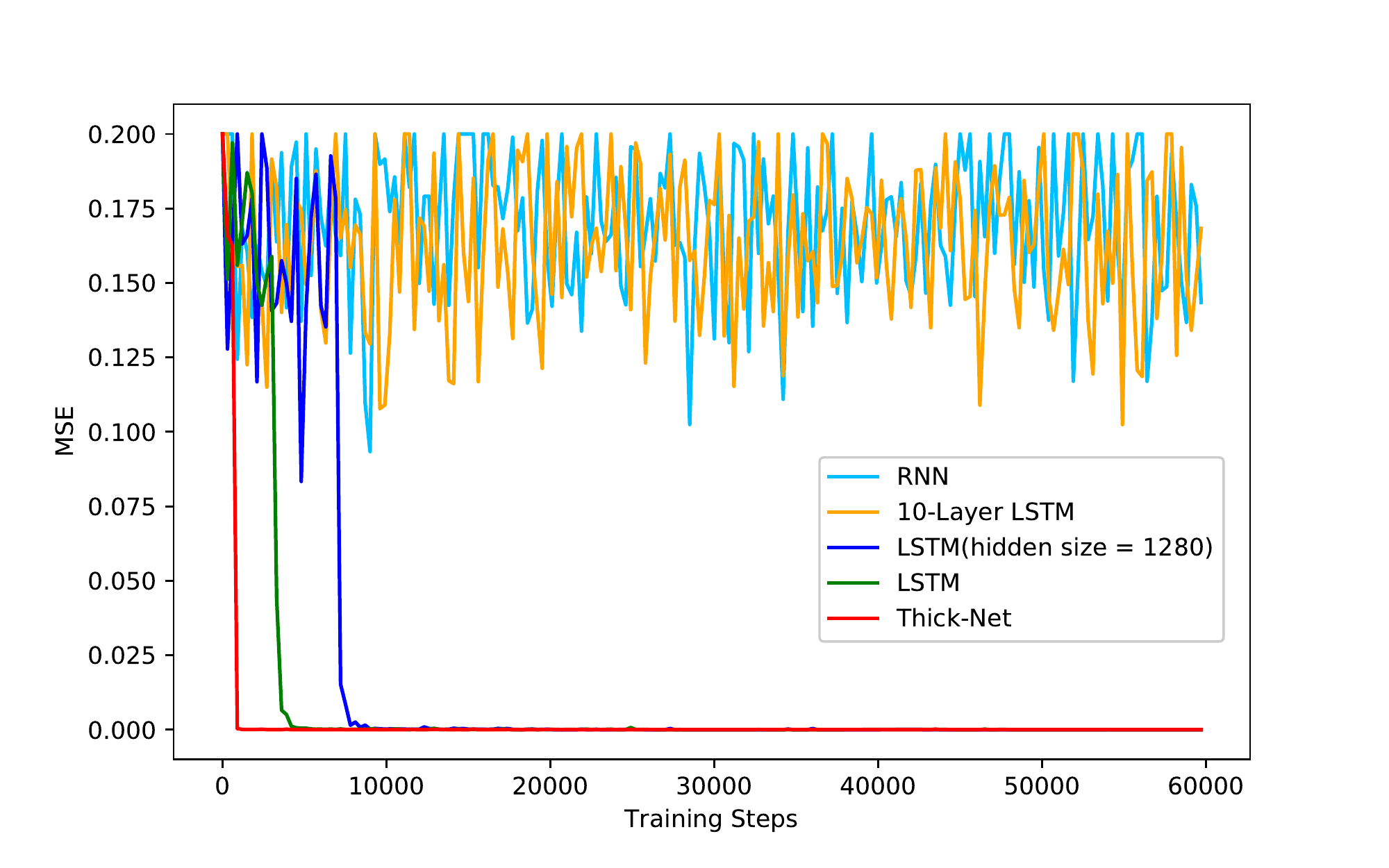}
\label{fig:T=100}
\end{minipage}%
}%
\subfigure[T=500]{
\begin{minipage}[t]{0.25\linewidth}
\centering
\includegraphics[width=1.4in,trim=0 15 0 43,clip]{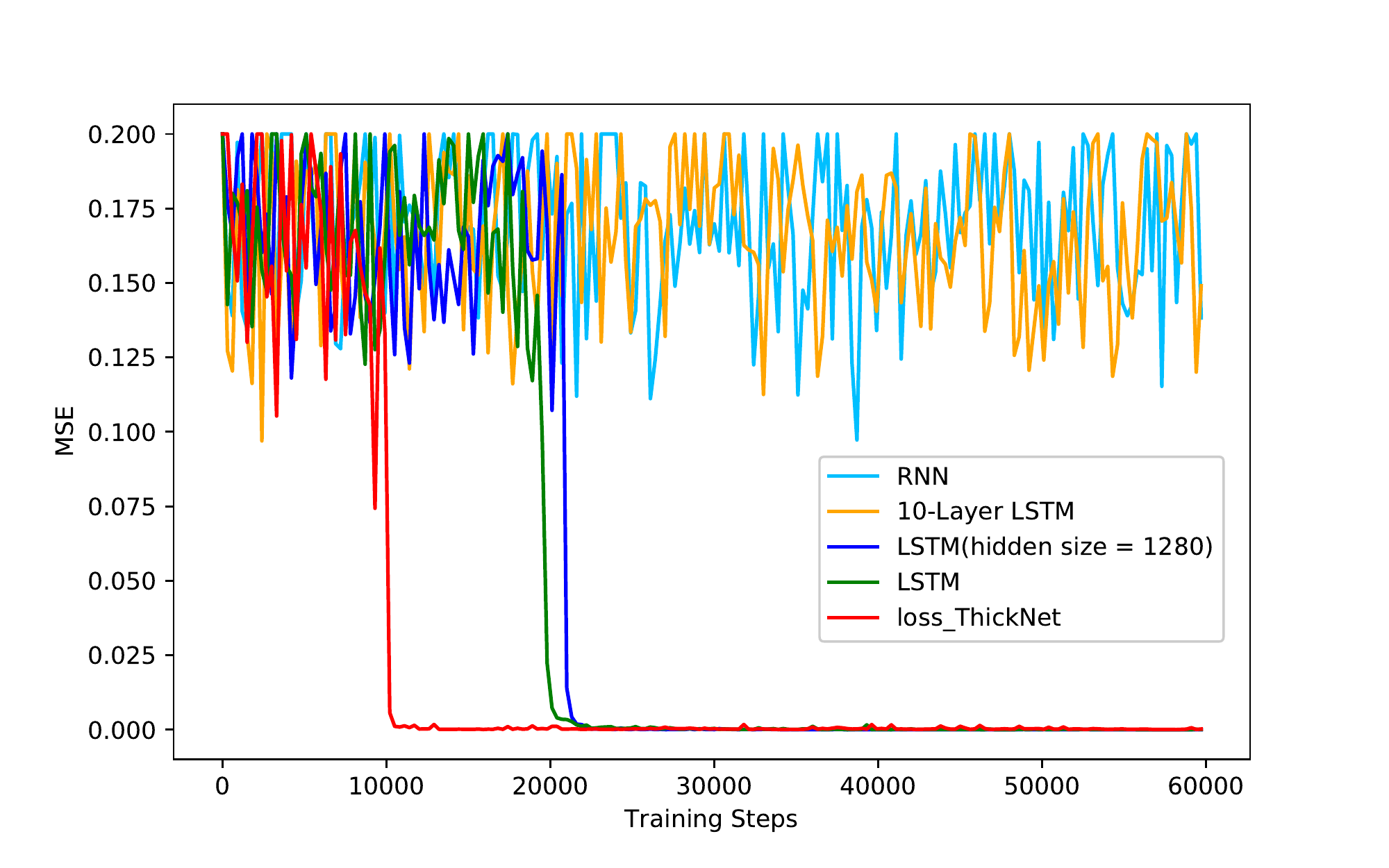}
\label{fig:T=500}
\end{minipage}%
}%

\centering
\caption{The figure (a) and (b) illustrates how loss function of MSE varies with the iterations of training. x-axis represents the training step and y-axis represents the loss function of MSE. (a)Results of Thick-Net with different operation functions. (b)Results of Thick-Net with different thickness $n$. The figure (c) and (d) shows comparison between Thick-Net and baseline models with different vector length $T$. (c)$T=100$ (d)$T=500$.}
\label{AddingProblem-text}
\end{figure}

%\begin{figure}[t]
%\vskip 0.2in
%\begin{center}
%\subfigure[Different Operation Functions]{
%\centerline{\includegraphics[width=\columnwidth,trim=0 15 0 43,clip]{ThickNet_pool.pdf}}
%\label{fig:Different Operation Functions}}
%\quad

%\subfigure[Different Thickness]{
%\centerline{\includegraphics[width=\columnwidth,trim=0 15 0 43,clip]{ThickNet_n.pdf}}
%\label{fig:Different Thickness}}
%\caption{The figure illustrates how loss function of MSE varies with the iterations of training. x-axis represents the training step and y-axis represents the loss function of MSE. (a)Results of Thick-Net with different operation functions. (b)Results of Thick-Net with different thickness $n$.}
%\label{AddingProblem-text}
%\end{center}
%\vskip -0.2in
%\end{figure}

The Addition Problem \cite{conf/icml/ArjovskySB16} is a basic simulation task for evaluating RNN models. Two vectors of length T are taken as input. The first vector is uniformly sampled over the range $(0; 1)$, while the second vector consists of two entries being $1$ with the remainder being $0$. The final output is the dot product of two vectors. The lengths of vectors are as two different values, i.e., T= 100 and 500.

%\begin{figure}[t]
%\begin{center}
%  \centerline{\includegraphics[width=0.85\columnwidth,trim=0 25 0 25,clip]{ThickNet2.pdf}}
%\caption{This figure shows comparison between Thick-Net and baseline models with different vector length $T$. (a)$T=100$ (b)$T=500$. }
%\label{Adding-problem}
%\end{center}
%\vskip -0.2in
%\end{figure}

%\begin{figure}[t]
%\centering\subfigure[T=100]{
%\begin{minipage}[t]{0.5\linewidth}
%\centering
%\includegraphics[width=3in,trim=0 15 0 43,clip]{T=100.pdf}
%\label{fig:T=100}
%\end{minipage}%
%}%
%\subfigure[T=500]{
%\begin{minipage}[t]{0.5\linewidth}
%\centering
%\includegraphics[width=3in,trim=0 15 0 43,clip]{T=500.pdf}
%\label{fig:T=500}
%\end{minipage}%
%}%
%
%
%\centering
%\caption{This figure shows comparison between Thick-Net and baseline models with different vector length $T$. (a)$T=100$ (b)$T=500$.}
%\label{AddingProblem-text}
%\end{figure}

When dealing with the adding problem, the features extracted through the maximization operation is essential. In addition to maximization operation, operation can also apply other functions, such as choosing the average or random values. To test the effectiveness of choosing the maximum, we draw a comparison among these three functions. And the Figure~\ref{fig:Different Operation Functions} explicitly demonstrates that the LSTM using maximization operation converges the fastest among these three functions.

The value of thickness $n$ has also been discussed in this part. The thickness $n$ is chosen to be $2, 4, 8, 16$ during the trial. From the Figure~\ref{fig:Different Thickness}, Thick-Net can converge faster with thickness $n=4$, and thereby being more efficient and easier to optimization.

One-layer Traditional RNN and LSTM with hidden sizes of $128$ are used as baseline models for experiments. The proposed Thick-Net applies a one-layer neural network with the same size and thickness $n$ of $4$. In order to draw a more comprehensive comparison, a deeper network(LSTM with $10$ layers, hidden size of $128$) and a wider network(LSTM with $1$ layer and hidden size of $1280$) are included in the experiment separately.

Mean squared error (MSE) is used as the objective function and the Adam optimization method \cite{kingmaadam} is used for training. The initial learning rate is set to $2\times 10^{-3}$. The training data and testing data are all generated randomly throughout the experiments.

The results compared with baseline models are shown in Figure~\ref{fig:T=100} and ~\ref{fig:T=500}. For short sequences (T = 100), the LSTM performs well and the proposed Thick-Net can converge to a very small error even more quickly. Unlike Thick-Net, the increase of the width of network will slow down the convergence. And a deeper network, as well as the traditional RNN, fails to minimize the error anymore. As the length of vectors increases (T = 500), traditional RNN and 10-layer LSTM still cannot converge to the minimum error. The convergence of Thick-Net is relatively quick compared with traditional LSTM and wider LSTM with hidden size of 1280.

Figuratively, the increase in the width and depth of network cannot always improve the performance in this task. In terms of wider network, the growth in hypothesis space cannot enhance the generalization ability. While for deeper network, the repetitive learning process of previous hypothesis space is not able to minimize the error. The proposed Thick-Net neither amplifies the hypothesis nor increases the numbers of nodes in depth's respect. Instead, it increases the number of parameters within each node. In this way, the generalization ability and the optimization rate can be significantly increased in this experiment.
\subsection{Permuted Sequential MNIST}
We evaluate our structure on sequential MNIST classification task \cite{journals/corr/LeJH15}. The model processes each image one pixel at each time step and finally predicts the label. The permuted MNIST (pMNIST) is also considered which makes the task harder. In pMNIST, the pixels are processed in a fixed random order.

\begin{table}[h]

\vskip 0.15in
\begin{center}
\begin{small}
\begin{sc}
\begin{tabular}{lccr}
\toprule
%Data set & Naive & Flexible & Better? \\
&MNIST   &pMNIST\\ %\hline
\midrule
IRNN  & 95.0\% & 82\% \\
%uRNN    & 95.1\% & 91.4\% \\
%RNN-path   & 96.9\% & - \\
LSTM & 98.2\% & 88.0\% \\
LSTM+Recurrent dropout    & - & 92.5\% \\
LSTM+Recurrent batchnorm   & - & 95.4\% \\
LSTM+Zoneout   & - & 93.1\% \\

\hline
Thick-Net(one-layer)   & 98.6\% & 96.0\% \\
\bottomrule
\end{tabular}
\caption{Results for the sequential MNIST and permuted MNIST}
\label{pMNIST-table}
\end{sc}
\end{small}
\end{center}
\vskip -0.1in
\end{table}

Our baseline contains traditional RNN and LSTM, a LSTM with batch-normalizing transformation \cite{conf/icml/LiHT0QWL18} and a LSTM adding zoneout \cite{conf/coling/SemeniutaSB16} on the recurrent connections. Each model has one layer of 100 hidden units. Our proposed Thick-Net applies a one-layer network with the same size and thickness $n$ of 10. Stochastic gradient descent on minibatches of size 128, with gradient clipping at 1.0 and step rule determined by Adam with learning rate $2\times 10^{-3}$.
%\begin{figure}[h]
%\vskip 0.2in
%\begin{center}
%\subfigure[pMNIST]{
%\centerline{\includegraphics[width=\columnwidth,trim=0 10 0 14,clip]{pMNIST}}
%\label{fig:pMNIST}}
%\quad
%\subfigure[Text Classification]{
%\centerline{\includegraphics[width=\columnwidth,trim=0 5 0 30,clip]{TextClass}}
%\label{fig:TextClass}}
%\caption{Results for permuted MNIST and Text Classification tasks}
%\label{pMNIST-text}
%\end{center}
%\vskip -0.2in
%\end{figure}

\begin{figure}[t]
\centering
\subfigure[pMNIST]{
\begin{minipage}[t]{0.5\linewidth}
\centering
\includegraphics[width=3in,trim=0 10 0 14,clip]{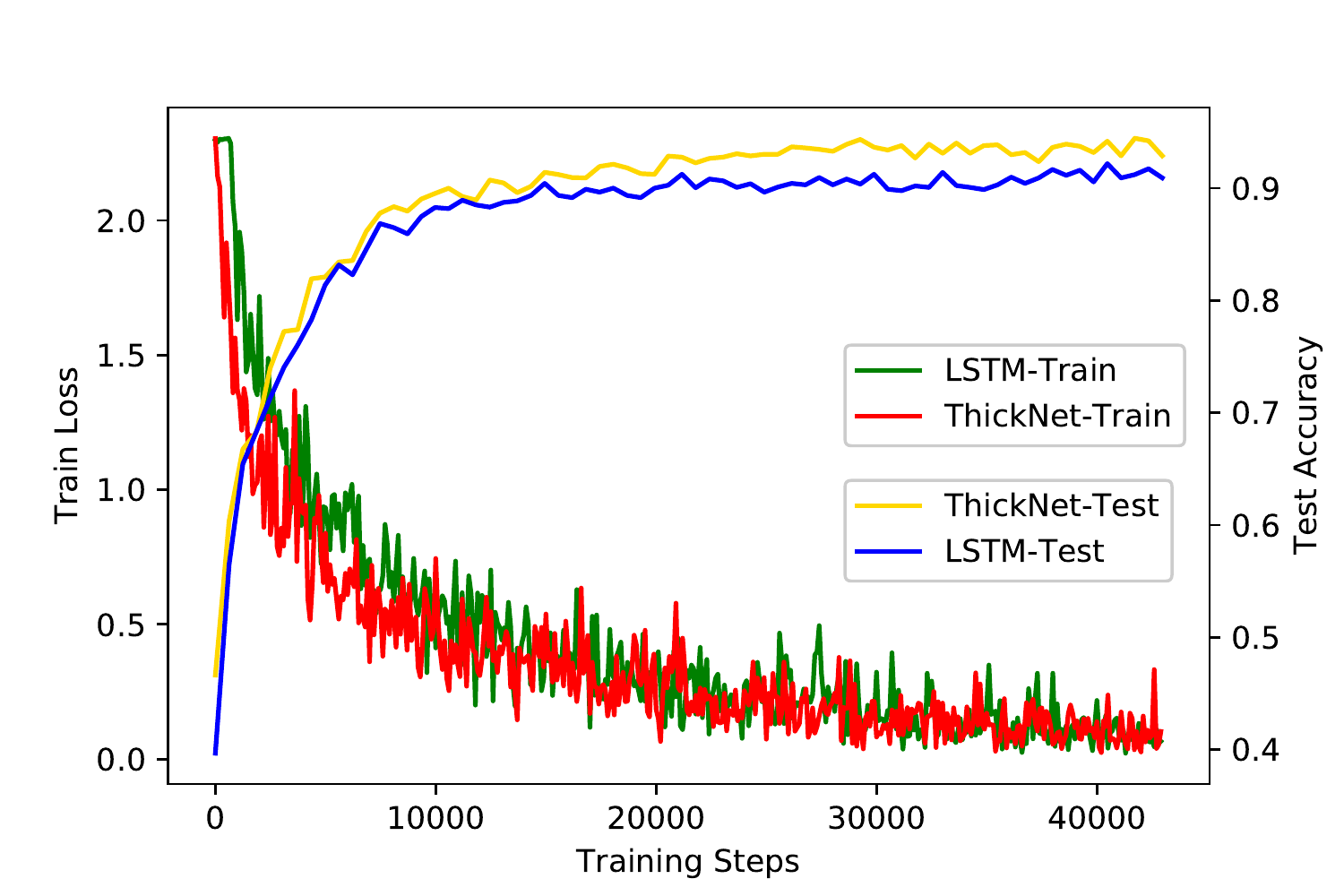}
\label{fig:pMNIST}
\end{minipage}%
}%
\subfigure[T=500]{
\begin{minipage}[t]{0.5\linewidth}
\centering
\includegraphics[width=3in,trim=0 5 0 30,clip]{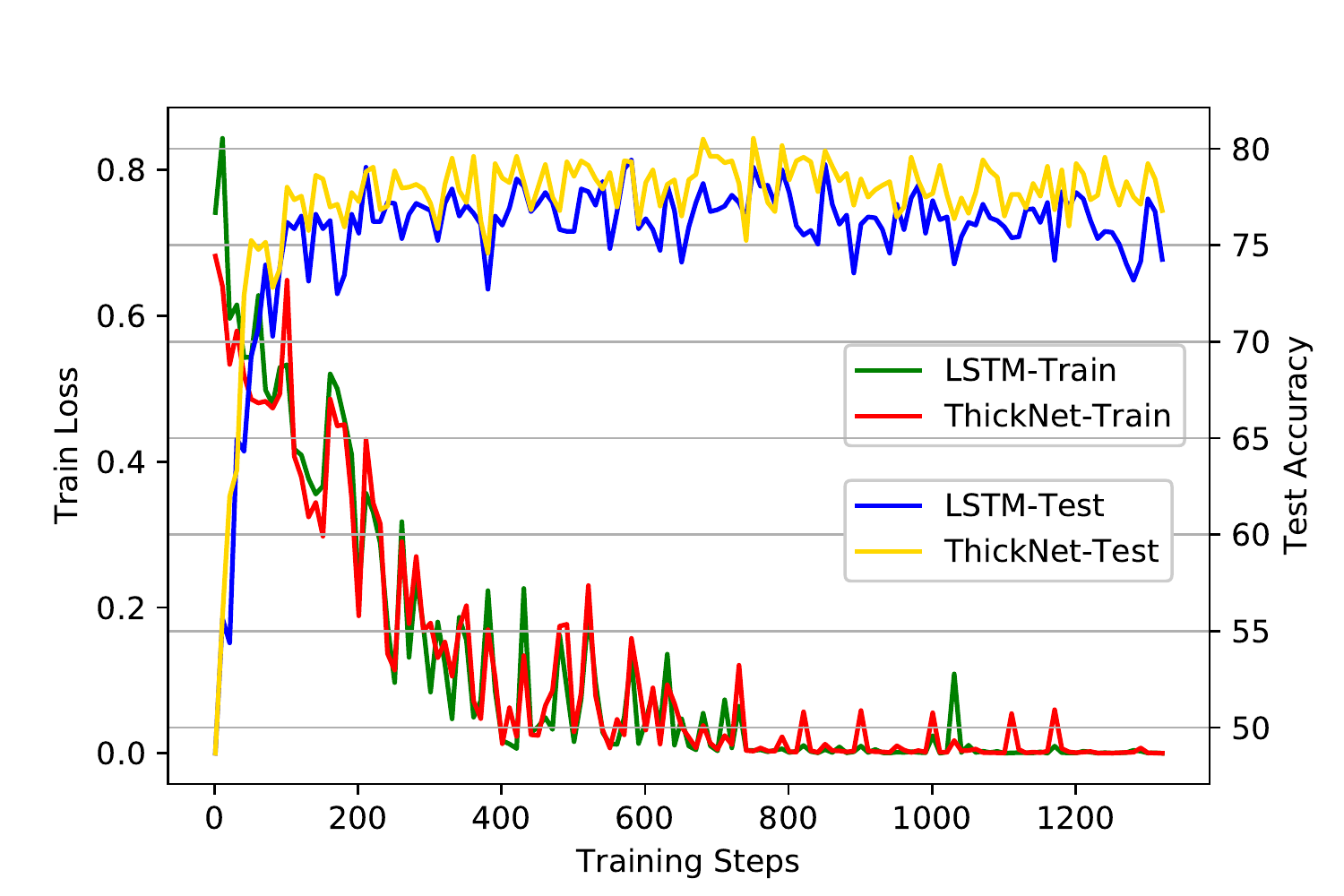}
\label{fig:TextClass}
\end{minipage}%
}%
\centering
\caption{Results for permuted MNIST and Text Classification tasks}
\label{pMNIST-text}
\end{figure}

The results of accuracy are demonstrated in Table~\ref{pMNIST-table} for comparison with the baseline models.
The results shown in Figure~\ref{fig:pMNIST} report the value of loss function for training data and accuracy rate for test data to evaluate the results obtained from Thick-net and traditional LSTM. From the figure, the accuracy rate reaches the state-of-the-art level while our model converges to the results earlier than the LSTM. The graph of loss function shows that our proposed Thick-Net outperforms the traditional LSTM due to (1) faster optimization rate and (2) better performance on handling overfitting problems.

\subsection{Text Classification}
In this section, we evaluate our proposed Thick-Net on text classification tasks.
We test our model on three datasets of classic sentence classification tasks.
\begin{itemize}
\item MR: Movie reviews\cite{conf/acl/PangL05} with two classes(positive and negative).
\item Subj: Subjectivity dataset\cite{conf/acl/PangL04}. Each sentence has a subjective or objective label.
\item TREC: Question classification dataset\cite{conf/coling/LiR02} contains 6 different question types for classification.
\end{itemize}

\begin{table}[t]
\caption{Comparisons of the accuracy rate between our Thick-Net and baselines on three text classification benchmarks.}
\centering
\label{TextClass-table}
\begin{tabular}{lcccr}
\toprule
%Data set & Naive & Flexible & Better? \\
&MR             &Subj         &TREC\\
\midrule
LSTM \cite{Bahdanau2016}                          & 75.9\%        & 89.3\%     & 86.8\%  \\
BiLSTM \cite{Bahdanau2016}             & 79.3\%        & 90.5\%     & 89.6\%  \\
Tree-LSTM \cite{conf/acl/TaiSM15}                    & 80.7\%        & 91.3\%     & 91.8\%  \\
LR-LSTM \cite{Qian2017Linguistically}                   & 81.5\%        & 89.9\%     & -       \\
%HS-LSTM                        & 82.1\%        & 93.7\%     & -       \\
\hline
Thick-Net(one-layer)                         & 80.63\%           & 93.9\%      & 92.0\%  \\
\bottomrule
\end{tabular}
\end{table}

We implement our model using the method Adam-an algorithm for first-order gradient-based optimization of stochastic objective functions. In this task, we set the learning rate as $0.01$, the dropout rate as $0.5$ and we also apply cross-entropy loss function to evaluate our results.

The results of accuracy are demonstrated in Table~\ref{TextClass-table} for comparison with the baseline models.

\subsection{Language Modeling}
Language modeling (LM) task is to build the essential statistical model that can capture how meaningful sentences can be constructed from individual words, and then use this trained model to predict the next word.

We test our model over the Penn Treebank (PTB) \cite{marcus93building}. The PTB data set has been considered as a central data set in language modeling task, and it does not contain capital letters, numbers or punctuation. The vocabulary list contains $10000$ unique words.

In this experiment, we choose different baseline methods to evaluate and compare with our structure including: LSTM, Variational LSTM \cite{conf/nips/GalG16}, Pointer Sentinel-LSTM \cite{conf/iclr/MerityX0S17}, LSTM + continuous cache pointer \cite{conf/iclr/GraveJU17}, Variational LSTM + augmented loss \cite{conf/iclr/InanKS17}, Variational RHN cite\cite{conf/interspeech/PundakS17}, 4-layer skip connection LSTM \cite{conf/iclr/MelisDB18}, AWD-LSTM \cite{conf/iclr/MerityKS18}.

We implement our model using the NT-ASGD algorithm for training and use a batch size of $40$ for PTB. In our experiment, we set the initial learning rate as $30$, and we followed the practice in \cite{conf/iclr/MerityKS18} to set up the other initial parameters. To improve language modeling results, we run ASGD with $T=0$ and hot-started $w_0$ as a fine-tuning step, and pointer based attention models \cite{conf/iclr/MerityX0S17} have been applied in our model.

\begin{table}[t]
\caption{Results of word-level Penn Treebank for Thick-Net in comparison with results of baseline models, in terms of perplexity.}
\label{Penn-Treebank}
\centering
\begin{tabular}{lcccr}
\toprule
%Data set & Naive & Flexible & Better? \\
Model &Valid &Test\\ %\hline
\midrule
RNN \cite{article} & - & 124.7 \\
LSTM \cite{zaremba2014recurrent}&82.2 &78.4 \\
CharCNN \cite{Kim2016CharacterAwareNL}& -  &78.9 \\
Pointer Sentinel-LSTM \cite{conf/iclr/MerityX0S17}& 72.4& 70.9 \\
LSTM + continuous cache pointer \cite{Grave2016Improving}&-  &72.1 \\
Variational LSTM + augmented loss \cite{conf/iclr/InanKS17}&71.1 &68.5 \\
Variational RHN \cite{conf/icml/ZillySKS17}&67.9 &65.4 \\
4-layer skip connection LSTM \cite{conf/iclr/MelisDB18}&60.9 &58.3\\
AWD-LSTM (finetune+pointer, 3 layers) \cite{conf/iclr/MerityKS18} &53.9 &52.8 \\
\hline
Thick-Net(one-layer)   & 56.4 & 54.7 \\
Thick-Net(two-layer)   & 51.3 & 50.2 \\
\bottomrule
\end{tabular}
\end{table}

As shown in the Table~\ref{Penn-Treebank}, our model achieves the state-of-the-art level of performance on the sequential modeling task of language modeling using only one or two layers in neural networks. The result for two-layer Thick-Net in terms of perplexity of language model is better compared with other baselines. Thus, our proposed Thick-Net improves the accuracy of prediction by implementing thick nodes which can extract more prior features.
\section{Conclusion and Future Work}
In this paper, we have introduced a novel but simple architecture which can be provided flexibly in recurrent neural networks, named Thick-Net. Instead of width or depth, thickness , as another dimension, is increased to keep hidden state size unchanged. Unlike previous works, maximization operation is applied which can significantly strengthen the generalization capability of our proposed Thick-Net. Overall, the Thick-Net has three main contributions: achieving the state-of-the-art level of performance in accuracy, avoiding overfitting and easier optimization.

The selection of the maximum in each node is inspired by previous study on maxout units and max pooling, and rigorously proved by the experimental results. Thus, in the future work, we will explore our Thick-Net by implementing the attention mechanism to more precisely decide which value should be selected in each node. Additionally, more recurrent networks can benefit from the Thick-Net on sequential learning tasks.
%% The file named.bst is a bibliography style file for BibTeX 0.99c
\bibliographystyle{elsarticle-num}

\end{document}